\documentclass[runningheads,a4paper]{llncs}

\usepackage{amssymb}
\usepackage{amssymb}
\usepackage{amsmath}
\usepackage{mathtools}
\usepackage{multirow}
\usepackage{subfig}
\usepackage{bm}
\usepackage{color, soul}
\usepackage[font={footnotesize}]{caption}
\usepackage{array}
\usepackage{paralist}
\usepackage[dvipsnames]{xcolor}
\usepackage[inline]{enumitem}
\usepackage{url}
\usepackage[draft]{pgf}
\usepackage{todonotes}

\usepackage{times}
\usepackage{epsfig}
\usepackage{graphicx}
\usepackage{amsmath}
\usepackage{amssymb}
\usepackage{multirow}
\usepackage{subfig}
\usepackage{hyperref}
\usepackage{xcolor}
\usepackage[font={footnotesize}]{caption}

\urldef{\mail}\path|asaggese@unisa.it,n.strisciuglio@rug.nl |
\newcommand{\keywords}[1]{\par\addvspace\baselineskip
\noindent\keywordname\enspace\ignorespaces#1}
  
\newcolumntype{C}[1]{>{\centering\let\newline\\\arraybackslash\hspace{0pt}}m{#1}}

\DeclareSymbolFont{extraup}{U}{zavm}{m}{n}
\DeclareMathSymbol{\vardiamond}{\mathalpha}{extraup}{87}

\begin{document}
\sloppy
\mainmatter  

\title{Action recognition by learning pose representations}

\titlerunning{Lecture Notes in Computer Science: Authors' Instructions}

%

\author{Alessia Saggese\inst{1} \and Nicola Strisciuglio\inst{2}
 \and Mario Vento\inst{1} \and Nicolai Petkov\inst{2}}
\authorrunning{A. Saggese et al.}

\institute{
Dept. of Information Eng., Electrical Eng. and Applied Mathematics (DIEM), \\
University of Salerno, Italy
\and
Johann Bernoulli Institute for Mathematics and Computer Science,\\
University of Groningen, The Netherlands\\
\mail}

%
%

\toctitle{Lecture Notes in Computer Science}
\tocauthor{Authors' Instructions}
\maketitle

\begin{abstract}
Pose detection is one of the fundamental steps for the recognition of human actions.
In this paper we propose a novel trainable detector for recognizing human poses based on the analysis of the skeleton. 
The main idea is that a skeleton pose can be described by the spatial arrangements of its joints.
Starting from this consideration, we propose a trainable pose detector, that can be configured on a prototype skeleton in an automatic configuration process. The result of the configuration is a model of the position of the joints in the concerned skeleton. In the application phase, the joint positions contained in the model are compared with the ones of their homologous joints in the skeleton under test. The similarity of two skeletons is computed as a combination of the position scores achieved by homologous joints.
In this paper we describe an action classification method based on the use of the proposed trainable detectors to extract features from the skeletons. We performed experiments on the publicly available MSDRA data set and the achieved results confirm the effectiveness of the proposed approach.

\end{abstract}

\graphicspath{ {./figures/}
			{./figures/configuration/}
			{./figures/application/}
			{./figures/dataset/}}		

\section{Introduction}
Automatic estimation of poses and recognition of actions are of great interest for the community of researchers in computer vision because of their various applications: think, as an example, to assisted living in robotics or to intelligent surveillance systems, where action recognition is a key step for the evaluation and the classification of people behaviors \cite{avss_deep}. 

Comprehensive reviews of methods for recognition of human poses and actions were published in the past years~\cite{Poppe10,Aggarwal11}. Although it is not possible to define a strict partition for the existing methods, typically the two categories are considered, depending on the typology of  the representation of the human pose or of the human action. Indeed, there are methods based on \emph{local} descriptors and methods based on\emph{global} descriptors. 


In the case of methods based on local descriptors, image patches are analyzed in a bottom-up approach. Usually, the descriptors computed on these local patches are combined in a hierarchical architecture, e.g. the bag of features approach~\cite{dollar05,Kovashka10,Lee14}.
One of the main advantages deriving from the use of local descriptors is that they can be computed on the whole image, without the need to perform preliminary detection and tracking of the objects. This approach can be also employed in cases where the scenes are crowded. However, this advantage is typically paid in terms of high effort required for the extraction of the descriptors. Furthermore, the accuracy of  methods based on local descriptors strongly depends on the amount as well as on the reliability of the interest points detected by the algorithms.

%
In the second group, global descriptors are based on a top-down approach. The subjects of interest are detected and tracked by using traditional background subtraction and updating algorithms, and features derived from optical flow, skeletons, silhouettes or edges are extracted \cite{yan11,wang07,sba13,ofli12}.
In this case, only information related to the movement and to the pose of the human is taken into account. The higher the accuracy in detecting the human subject is, the better the performance of these approaches will be.
For these reasons, several methods based on global descriptors in the last years have been proposed.

One of the milestones in the research on action recognition was the work of Johansson \emph{et al.}~\cite{Johansson73}, that demonstrated that several human actions can be recognized by only looking at the position of the limb joints. In their experiments, these joints were marked by light-point sources and human observers could recognize actions by having no additional information.
Nowadays, several devices, such as the Microsoft Kinect and the MOCAP systems, allow to reliably extract the skeleton and the limb joint points of people moving in a scene. 
The skeleton is a set of rigid segments representing the bones, connected each other by joints that correspond to articulations. A particular configuration of the skeleton, expressed in terms of spatial arrangement of joints, can be considered a \emph{pose}. The temporal evolution of a set of consecutive poses can be associated to a specific \emph{action}. 
Within this framework, the theoretical and experimental studies conducted by Johansson \emph{et al.} are very important because demonstrated that the analysis of human skeleton is sufficient in many situations to recognize the poses and thus the actions performed by a human subject.

The importance of skeleton information has been further confirmed in a more recent experiment in~\cite{Yao11}, aimed to evaluate the discriminant power of different representations based on skeletons and more traditional low-level appearance features. Indeed, it has been demonstrated that pose features outperform low-level appearance features, even when skeletons highly corrupted by noise.
Furthermore, although the skeleton extraction typically requires a computationally-intensive pre-processing step, it is less sensitive to intra-class variations.
These are the main reasons why in the last years several methods for action recognition based on skeleton analysis have been proposed \cite{Vemulapalli14,Chaudhry13,Yang14,Dias16,Dias16_2}.

The problem of recognizing human actions by analyzing the skeleton can be approached at two levels: 
\begin{inparaenum}[\itshape a\upshape)]
\item given two skeleton poses, evaluate their similarity
\item given two sequences of poses, encode such sequences and evaluate their similarity.
\end{inparaenum}

In this paper we focus on the construction of  effective representations of the skeleton poses that we then employ as feature extractors in a method for recognizing human actions. Our main idea is based on the fact that a pose is described by a particular spatial arrangement of joint positions. We propose trainable features that learn a model of a given prototype skeleton pose in an automatic configuration process. The features that we propose can be used to evaluate the similarity of a skeleton pose to the ones used for training.

The concept of trainable features was originally introduced in~\cite{Azzopardi13}, where COSFIRE filters were proposed for object recognition and keypoint detection. 
The trainable character of COSFIRE features stands in the fact that their structure is not fixed in the implementation but it is rather learned in a automatic configuration process performed on given prototype samples.  
The automatic configuration of features avoids to design a set of hand-crafted features to transform the raw data into a suitable representation or feature vector to be used in combination with a classifier system.
This is a kind of \emph{representation learning}, where the important characteristics of the patterns of interest are directly learned from training samples. 
Trainable features derived from COSFIRE have been successfully applied to various problems in image processing, such as contour detection~\cite{AzzopardiPetkovCORF2012}, delineation of blood vessels in medical images~\cite{AzzopardiMEDIA2015,StrisciuglioVIP15}, color-object detection~\cite{Gecer2017} and adapted to audio event detection~\cite{StrisciuglioCOPE2016}.	

In this paper we propose a trainable pose detector that is automatically configured by modeling the position of the joints with respect to a reference point in a given prototype skeleton (we consider the body barycenter as reference point). 
In the application phase, we compute the response of a pose detector by combining a score value that is computed for each joint in the model. The score of a joint indicates the similarity of the position of such joint with the one of its homologous in the prototype skeleton.
The proposed pose detector introduces tolerance in the position of each joint  in order to account for deformations of the skeleton with respect to the ones configured in the model.

In the proposed method, we configure a number of pose detectors from training samples and use them as feature extractors to construct a feature vector that describes the pose of a skeleton. The so constructed vectors can be, then, used to train any possible classifier. In our experiments, we employed a multi-class Support Vector Machine (SVM).
We carried out experiments on the publicly available MSRDA data set and obtain comparable results with the ones achieved by existing methods based on skeleton pose analysis.

The paper is organized as follows: in Section~\ref{sec:proposed} we provide details about the proposed pose feature detector and the classification method; in Section~\ref{sec:exp} we report about the experimental results that we achieved and compare them with the ones obtained by existing methods, while we draw conclusions in Section~\ref{sec:concl}.


\section{Method}
\label{sec:proposed}

The basic idea underlying the proposed approach is to automatically learn a representation of skeleton poses by modeling the spatial arrangement of skeleton joints with respect to a reference point (in this work we consider the barycenter of the body). Two skeletons can be considered similar if the relative position of their homologous joints is similar. 
Hence, computing the similarity between two skeletons can be formulated as computing the similarity between the relative  position of their homologous joints. We propose trainable skeleton pose detectors, inspired by the principles of trainable COSFIRE filters for representation learning in pattern recognition. 
The proposed detectors are trained in an automatic configuration process performed on a given prototype skeleton and in the application face it is able to detect the same pattern and modified versions of it. In the following, we provide details of the configuration and application phases.


\subsection{Configuration}
Given a prototype skeleton $J$, the proposed pose detector learns a model of the position $(x_i,y_i)$  of its joints $j_i, i=1, \dots, n_J$ with respect to a reference point in a configuration process. The value $a_i$ is an identifier of the corresponding limb joint point to the skeleton joint point, while $n_J$ is the total number of joints in the prototype skeleton.

We construct a model $S$ of the prototype skeleton $J$ where each joint is described by  a $4$-tuple $(j_i, \rho_i, \phi_i, \omega_{i})$. The values  $\rho_i$ and $\phi_i$ are the polar coordinates of the position of the $i$-th joint with respect to the reference point $(x_b, y_b)$:
\begin{align}
\rho_i &= \sqrt{(x_i-x_b)^2 + (y_i-y_b)^2}\\
\phi_i &= \tan^{-1} \frac{y_i-y_b}{x_i-x_b}
\end{align}
The value $\omega_i$ is a parameter and represents the weight assigned to the joint as a measure of its importance in the concerned action. 

We divide the skeleton in two parts, corresponding to the upper and lower parts of the body, in order to increase the selectivity of the configured detectors for actions that involve only a specific part of the body. As an example, the action of waving hands involves only the upper part of the body, disregarding whether the subject is sitting or standing up. In such case, the weights of the joints in the lower body part are set to $0$, so that they do not contribute to the configuration phase and to the computation of the skeleton pose. This procedure helps to reduce the infra-class variations (e.g. waving hand for sat and stood up persons can be considered the same action although part of the skeleton pose is very different) and to increase the selectivity of the configured filters for specific poses. 

\subsection{Detection of skeleton similarity}
\begin{figure}[!t]
   \centering
    \setlength{\unitlength}{50mm}%
\subfloat[\label{fig:skeleton4} ]{
 	\includegraphics[height=\unitlength]{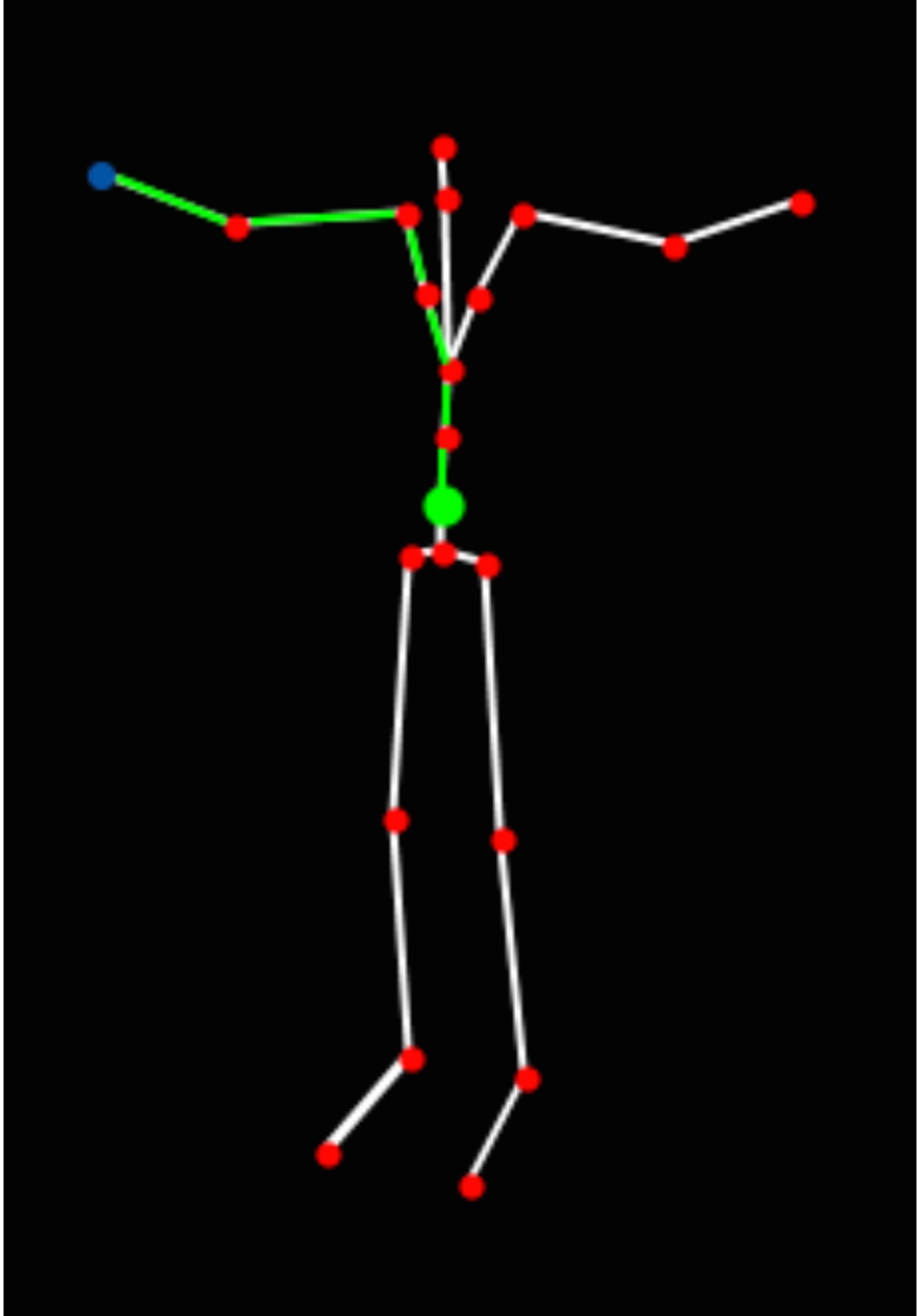}
} ~
\subfloat[\label{fig:skeleton2} ]{
 	\includegraphics[height=\unitlength]{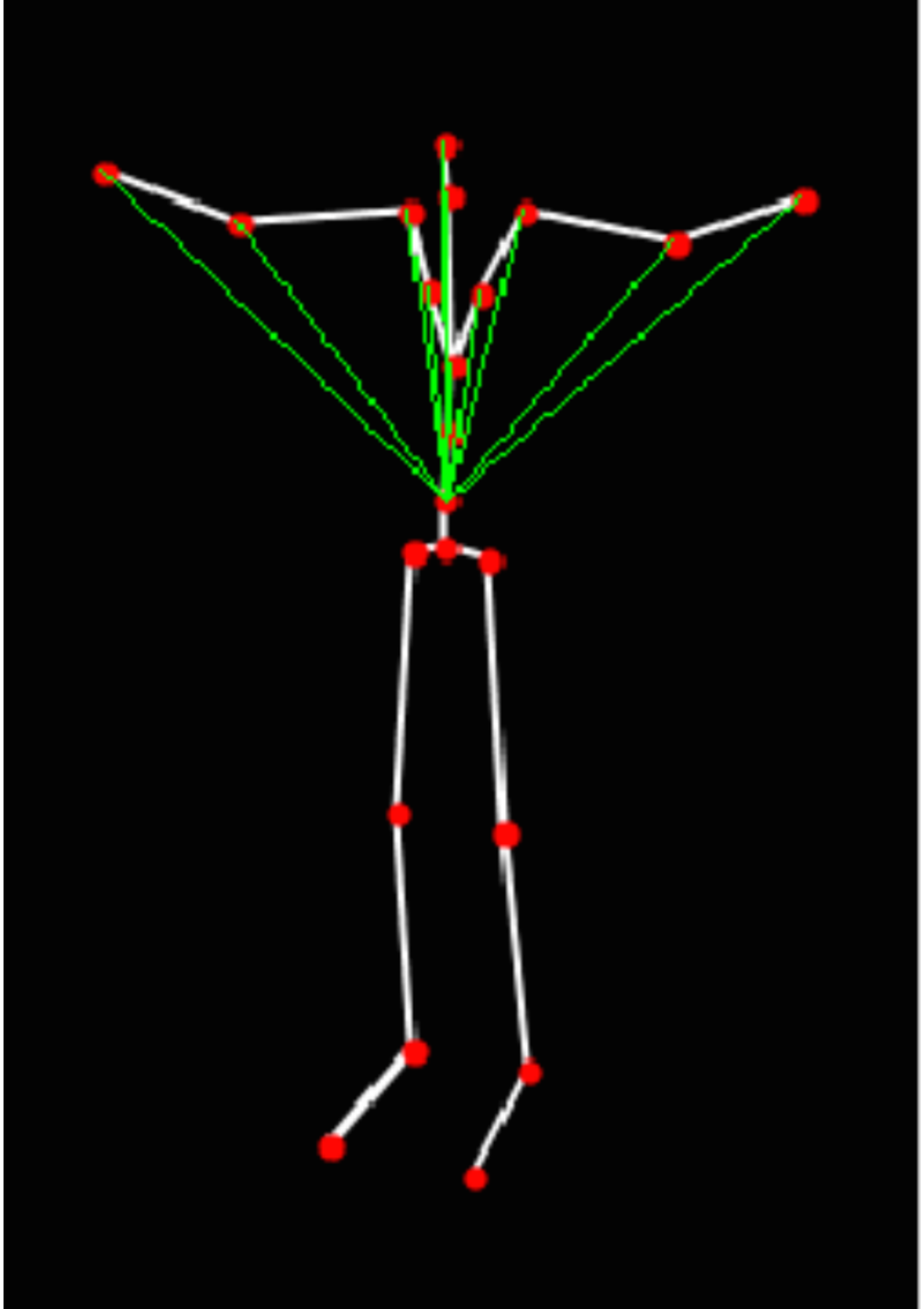}
}  ~
\subfloat[\label{fig:skeleton3} ]{
 	\includegraphics[height=\unitlength]{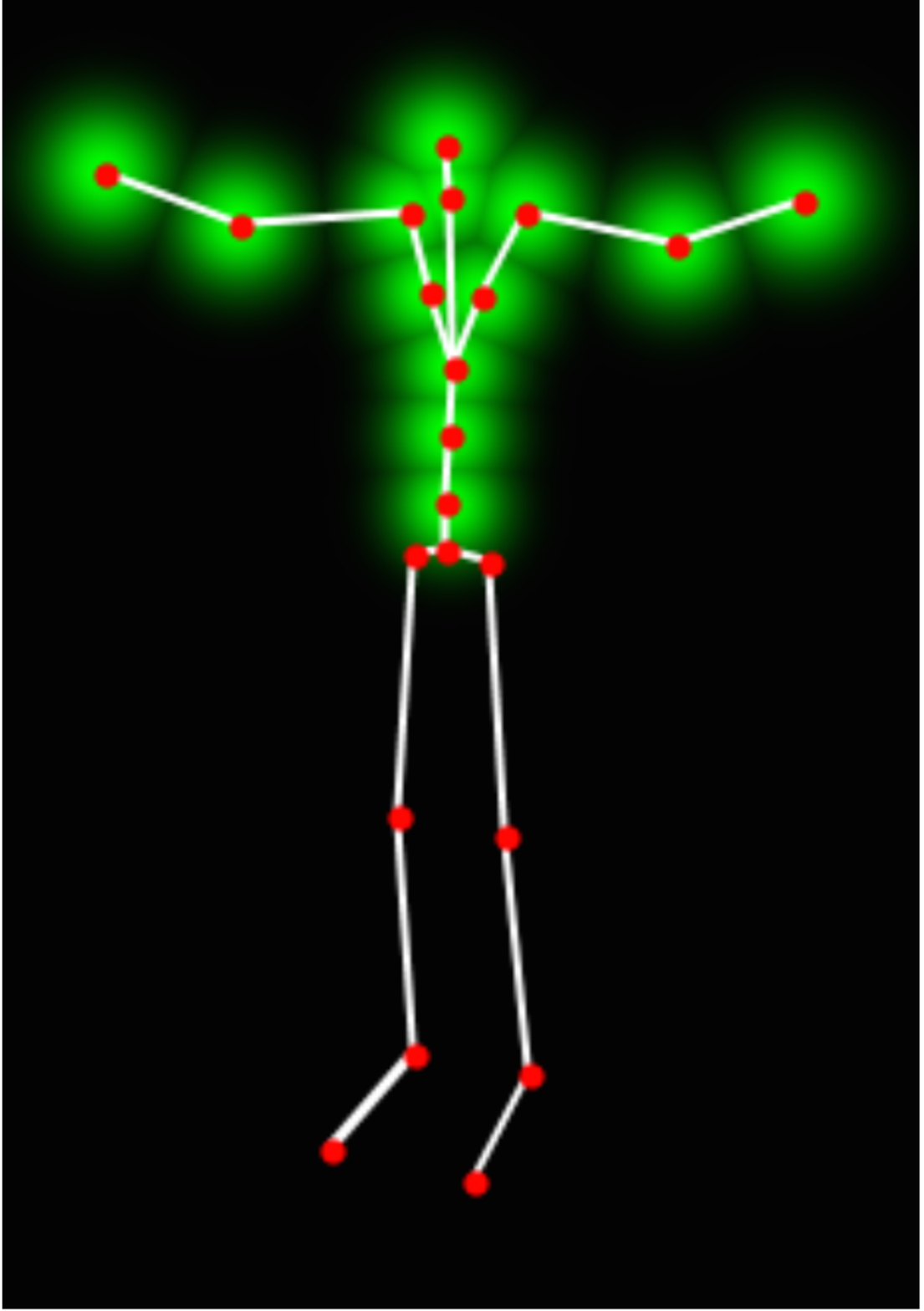}
} ~
   \caption{A prototype skeleton with examples of (a) skeletal distance between the reference point (green spot) and the joint that corresponds to the position of the right hand (in blue); the (b) considered joint positions (green lines) to configure a model of the upper part of the body; the (c) Gaussian weighting functions introduced to achieve tolerance to variations of the position of the joints in the configured model.}
   \label{fig:filter}
\end{figure}

We compute the similarity of a test skeleton to a prototype one by combining a score for each joint in the configured model. The value of the score of a joint depends on the position that it has in the test skeleton when compared to the position of the homologous joint in the model. 
In practice, we compare the distance relative to the reference point in the prototype and test skeletons, weighted with a Gaussian function, which allows for tolerance to spatial deformations.
Formally, the score $r(t_i, j_i)$ of the joint $t_i$ in the test skeleton with respect to its homologous $j_i$ in the prototype skeleton is computed as:
\begin{equation}
r(t_i,j_i) = 
e^{-\frac{D(t_i, j_i)}{2\sigma_i^2}}
\end{equation} 
where $D(t_i ,j_i)$ is the euclidean distance between the position of the joint $t_i$ in the test skeleton and of its homologous $j_i$ in the prototype skeleton. 
The value  $\sigma_i$ is the standard deviation of the Gaussian weighting function computed for the $i$-th joint. It is important to highlight that $\sigma_i$ regulates the tolerance to the position of the $i$-th joint with respect to the position of its homologous in the model $S$ determined in the configuration process.
The value of $\sigma_i$ as a linear function of the skeletal distance $\overline{d}(j_i, j_b)$ between the position of the reference point  $j_b$ and the one of the joint $j_i$:
\begin{equation}
 \sigma_i = \sigma_{0} + \alpha \cdot \overline{d}(j_i, j_b).
\end{equation}
The value of $\sigma_i$ increases with the skeletal distance of the $i$-th joint point from the reference point (i.e. the barycenter of the body). This determines a larger tolerance in the position of further joints and goes accordingly to the fact that terminal joints have more mobility than joints closer to the body.
The distance $\overline{d}$ is computed as the sum of the length of the segments that connect the joints $j_i$ and $j_b$ (see Fig.~\ref{fig:skeleton4}).
The parameters $\sigma_{0}$ and $\alpha$ regulate the amount of tolerance with respect to deformation from the prototype pattern in the application phase. 
The weighting function that we considered to account for tolerance in the position of the skeleton joints contributes to robustness to deformations of the prototype pattern. This property provides generalization capabilities to the proposed pose detector that  strongly responds to the same skeleton pattern used for configuration but also to similar (or deformed) versions of it.

We compute the similarity $R(T, S)$ between a test skeleton $T$ and a prototype skeleton $S$ as the combination of the measure of similarity between the positions of homologous joints in $T$ and $S$.
It is formally defined as the weighted geometric mean of the measure $r_i$ of the joint position similarity:
\begin{equation}
 R(T,S) = \bigg(\prod_{i=1}^{|S|} r(t_i,j_i)^{w_i}\bigg)^{1/\sum_{i=1}^{|S|}w_i}.
\end{equation}

\begin{figure}[!t]
   \centering
          \setlength{\unitlength}{50mm}%
\subfloat[\label{fig:application1} ]{
 	\includegraphics[height=\unitlength]{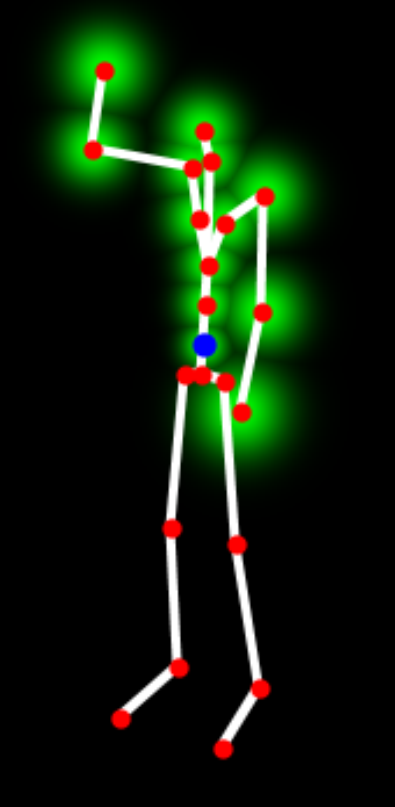}
}
~
\subfloat[\label{fig:application3} ]{
 	\includegraphics[height=\unitlength]{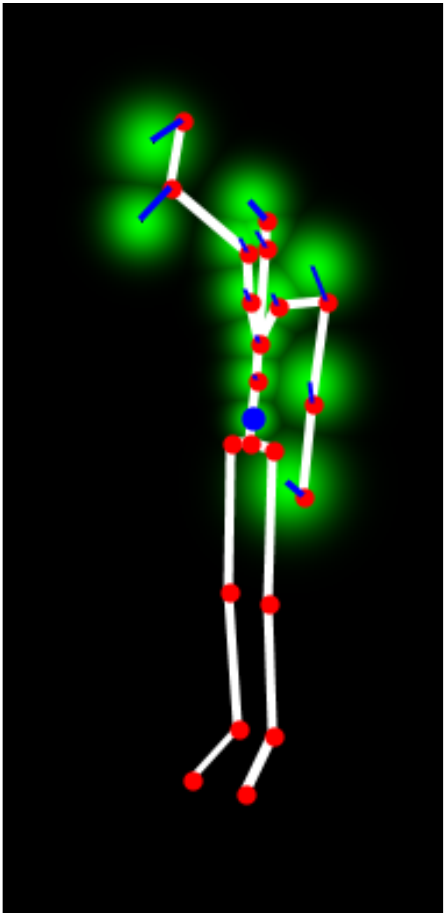}
} 
~
\subfloat[\label{fig:application2} ]{
 	\includegraphics[height=\unitlength]{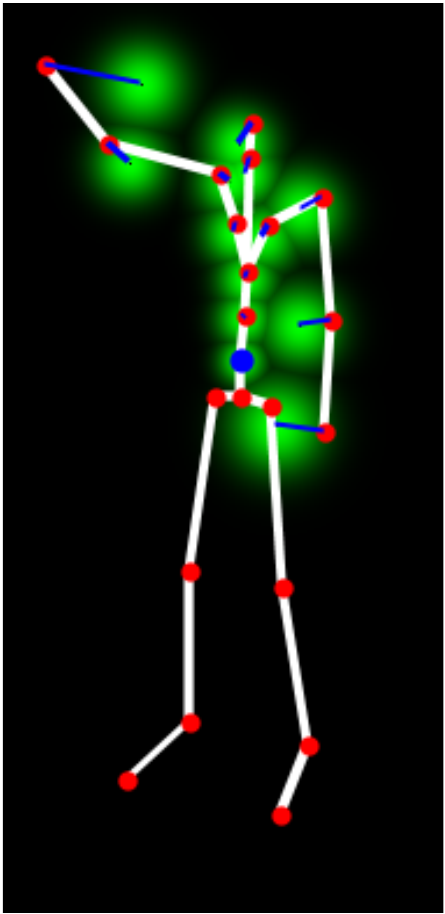}
} 
~
\subfloat[\label{fig:application4} ]{
 	\includegraphics[height=\unitlength]{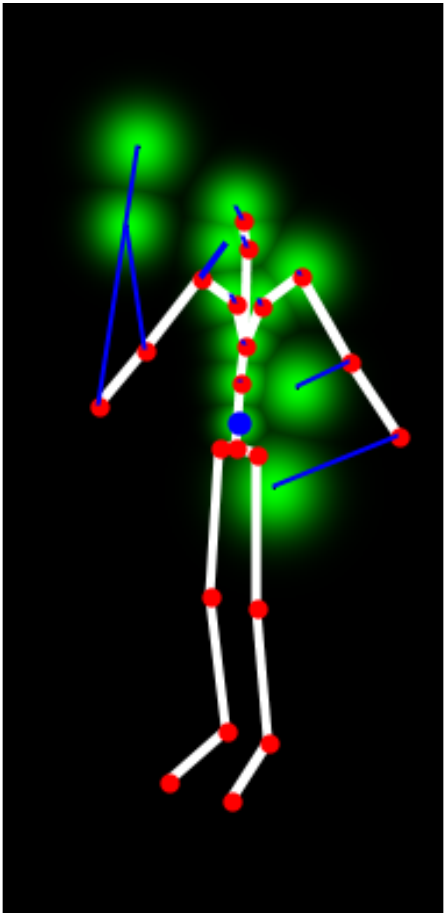}
} 
   \caption{A prototype filter (a) and the response to this filter of different skeletons (b,c,d), having response value $0.7$, $0.5$ and $0$ respectively. The blue lines represent the euclidean distance between the joints and the homologue prototype.}
   \label{fig:filter_application}
\end{figure}

A few examples of application of the proposed pose detector are shown in Fig.~\ref{fig:filter_application}, where the responses of the three skeletons in Fig.~\ref{fig:application3}, Fig.~\ref{fig:application2} and Fig.~\ref{fig:application4} with respect to the prototype skeleton in Fig.~\ref{fig:application1} are shown. Note that, as expected, the response achieved in the skeleton in Fig.~\ref{fig:application3}, is higher than the response achieved in Fig.~\ref{fig:application2} and in Fig.~\ref{fig:application1}, being $0.7$ vs $0.5$ and $0$ respectively.

\subsection{Robustness to reflection and scale}
\begin{figure}[!t]
   \centering
       \setlength{\unitlength}{42mm}%
   \subfloat[\label{fig:scale} ]{
 	\includegraphics[height=\unitlength]{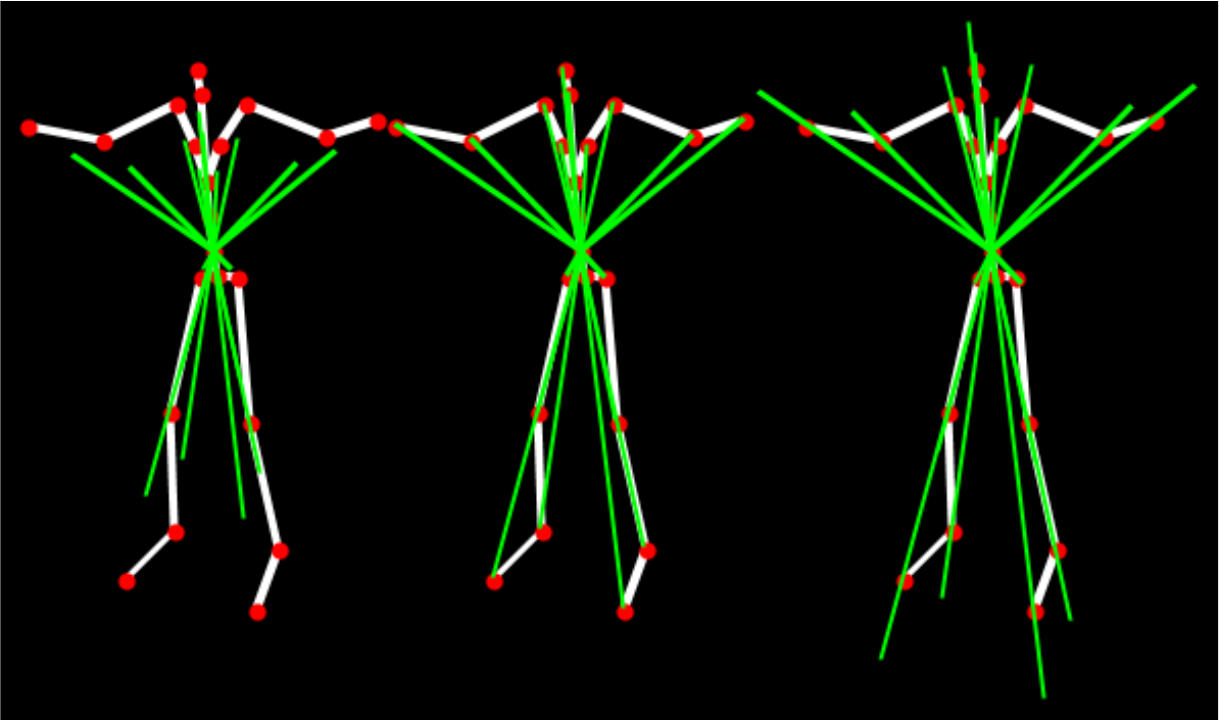}
} ~
\subfloat[\label{fig:reflection} ]{
 	\includegraphics[height=\unitlength]{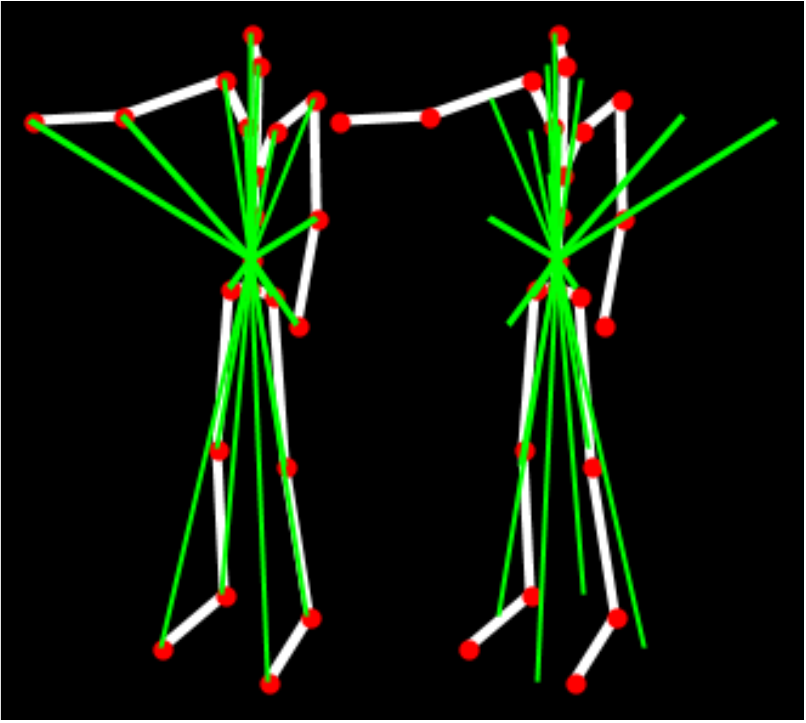}
}
   \caption{Examples of (a) scaled and (b) reflected versions the proposed pose detector. The green lines represent the positions of the transformed joint points in the scale- and reflection-tolerant versions of the detector.}
   \label{fig:scale_reflection}
\end{figure}

In order to achieve robustness with respect to scale and reflection transformations of the skeletons, we introduce and apply modified versions of the configured detectors. Examples are shown in Fig.~\ref{fig:scale_reflection}. 
For each joint $j_i$ in the model $S$, we also apply a reflected version $j_i^s$ (see Fig.\ref{fig:reflection}), by modifying the original tuples $(j_i, \rho_i, \phi_i, \omega_{i})$ as follows: 
\begin{equation}
(j_i^s, \rho_i, \pi - \phi_i, \omega_i),
\end{equation}
The modified angle value $\pi - \phi_i$ allows to identify those actions that can be performed in a symmetric way. As an example, a subject can perform the action of drinking both with the right and the left hands. The two actions are, however, symmetric with respect to the central axis of the body.

In order to also accounts for correct detection of actions performed at various distances from the camera, we apply a scaled version of the detector (see Fig.~\ref{fig:scale}) by modifying the tuples in the original models as follows:
\begin{equation}
(j_i, \eta \cdot \rho_i, \phi_i, \sigma_i, \omega_i).
\end{equation}
The factor $\eta$ introduces tolerance to actions that are performed at different distances from the camera. In our experiments, we set this value to $0.8$ (for actions performed at a closer distance) and $1.2$ (for actions performed at a further distance).

\subsection{Classification}
We configure a number $N$ of pose detectors for each of the $M$ actions of interest and use each configured detector as a skeleton feature extractor. We thus construct a feature vector of $M\times N$ elements, whose values are computed as the response of the configured detectors when applied to test skeletons.
The constructed feature vectors are used in combination with a classifier to associate a pose to a specific class of interest.

We employ a multi-class SVM classifier by combining the output of a pool of $M$ linear SVM classifiers, each one trained to recognize poses of a specific class of interest.
We train the $i$-th SVM using as positive examples the samples from the class $C_i$ and as negative examples  the samples from all the other classes (\emph{one-vs-all} scheme).  
In the classification step, each classifier assigns to the skeleton under test a score $s_i$ and we select the class $C_i$ that corresponds to the classifier with the highest score. In case all the classifiers compute a negative score we assign the skeleton under test to the background class, i.e. the pose is not recognized.

The so trained classifier is able to recognize skeleton poses in single frames, without taking into account their temporal evolution. In our experiments, we perform the classification of poses at level of single frames and, successively, aggregate the classification outputs to action-level. We classify an action by majority voting on the frame-level decisions.

\section{Experimental results}
\label{sec:exp}

\subsection{Data sets}
We performed experiments on a publicly available data set, namely the MSR Daily Activity 3D dataset (hereinafter MSRDA)~\cite{Wu12_dataset}.

\begin{figure}[!t]
   \centering
             \setlength{\unitlength}{35mm}
\subfloat[\label{fig:msrda2} ]{
 	\includegraphics[width=\unitlength]{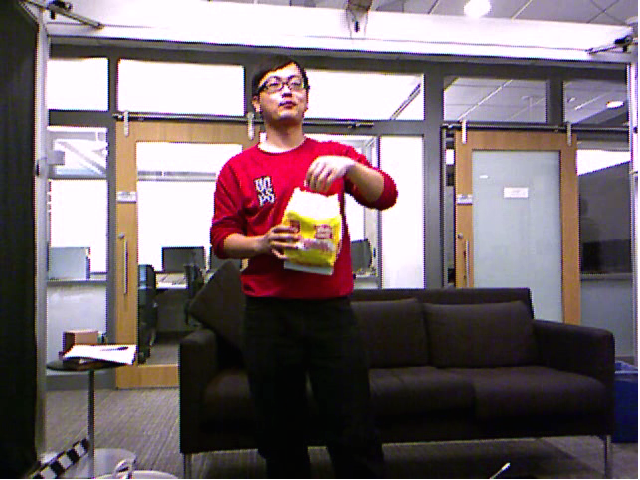}
} 
\subfloat[\label{fig:msrda2s} ]{
 	\includegraphics[width=\unitlength]{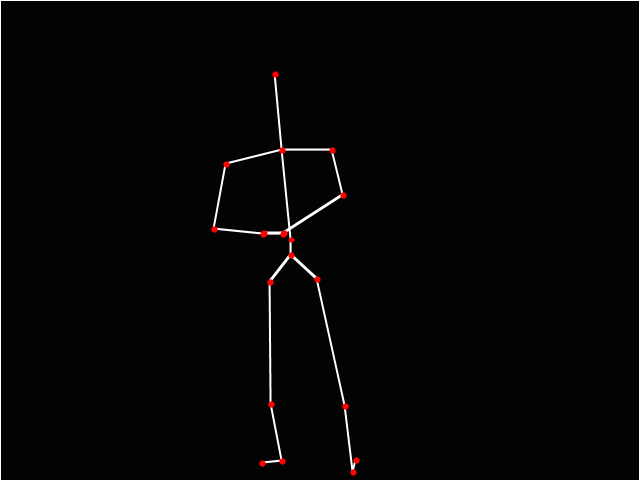}
} 
\subfloat[\label{fig:msrda_err1} ]{
 	\includegraphics[width=\unitlength]{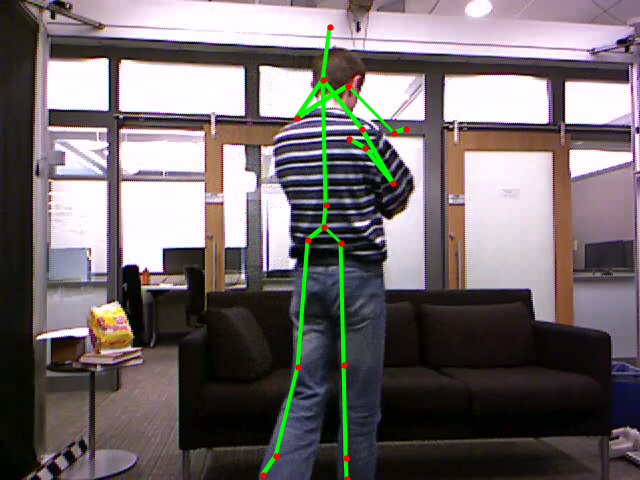}
}

\subfloat[\label{fig:msrda3} ]{
 	\includegraphics[width=\unitlength]{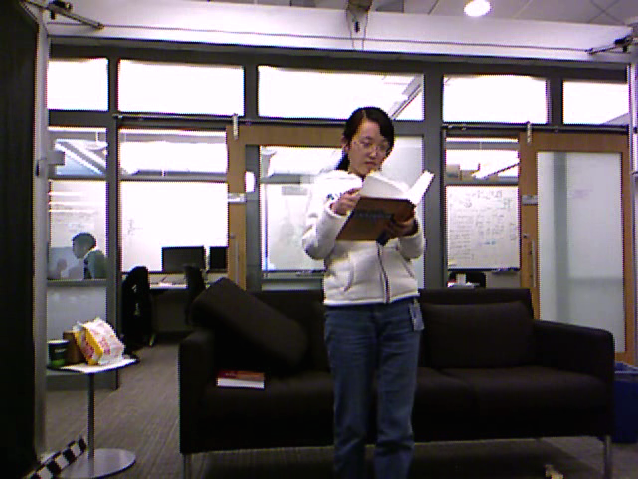}
} 
\subfloat[\label{fig:msrda3s} ]{
 	\includegraphics[width=\unitlength]{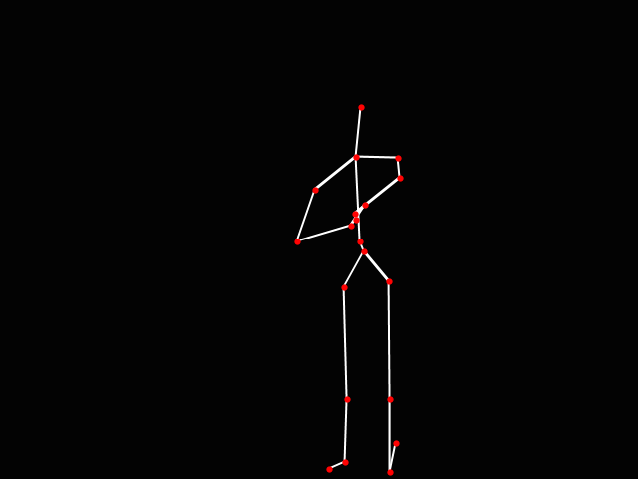}
} 
\subfloat[\label{fig:msrda_err2} ]{
 	\includegraphics[width=\unitlength]{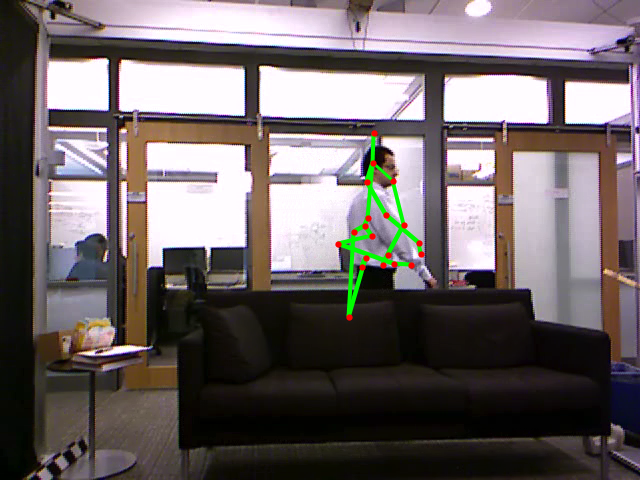}
} 

   \caption{Example frames extracted from the MSRDA data set from the action classes (a) Eat  and  (b) Read Book together with their (c,d) corresponding skeletons. Examples of (e,f) noisy skeletons.}
   \label{fig:dataset_msrda}
\end{figure}

The MSRDA data set is composed of $16$ classes of actions, which can be considered daily actions: drink (B1), eat (B2), read book (B3), call cellphone (B4), write on a paper (B5), use laptop (B6), use vacuum cleaner (B7), cheer up (B8), sit still (B9), toss paper (B10), play game (B11), lay down on sofa (B12), walk (B13), play guitar (B14), stand up (B15), sit down (B16). 
Each action is performed by $10$ people. In turn, each person repeats the actions twice, one while sitting on a sofa and the other while standing. A few example frames are shown in Fig.~\ref{fig:dataset_msrda}.
As stated by the data set authors, the position of the skeleton joints is  noisy, thus implying that several samples are corrupted and are difficult to be recognized by approaches base only on the analysis of the skeleton. We report two examples of noisy skeletons, over-imposed on the original images, in Fig.~\ref{fig:msrda_err1} and~\ref{fig:msrda_err2}.

\subsection{Results and discussion}

We evaluate the performance of the proposed method for action recognition by computing the average recognition rate, the error rate and the miss rate.  We compute the evaluation metrics by considering the classification outputs at action-level. The error and miss rates refer to samples that are classified to the wrong class and to the background class, respectively.

We achieved an average recognition rate of $64.0\%$ on the MSRDA data set, with an error rate of $34\%$ and a miss rate of $2\%$. We report detailed results for each class of actions in the MSRDA data set in Table~\ref{tab:miscl_msrda}. 
Classification errors are mainly due to mis-classification of the following action classes: \emph{read book} (B3), \emph{write a paper} (B5), \emph{use laptop} (B6) and \emph{play game} (B11). This is due to the fact that part of the poses that constitute these actions are in common, so making difficult to be distinguished by using only the skeleton information
The actions \emph{sit down} (B15), \emph{stand up} (B16), \emph{sit still} (B9) are also subjected to errors. In these cases, the absence of information related to the temporal sequence of poses does not contribute to correct decisions of the proposed approach.

\begin{table}
\begin{scriptsize}
\setlength\tabcolsep{0.2cm}
	\renewcommand{\arraystretch}{1.3}
	\centering
	\begin{tabular}{|c|c|c|c|c|}
			\hline
		\multicolumn{2}{|c|}{~} & \bfseries Recognition Rate & \bfseries Error Rate & \bfseries Miss  Rate \\ \hline
    			
			\multirow{17}{*}{\rotatebox{90}{\bfseries Classes}} & \textbf{B1} & ${0.70}$ &$0.2$ & $0.1$	 \\
			~ & \textbf{B2} & ${0.7}$ 	&$0.20 $ & $0.1$  \\
			~ & \textbf{B3} & ${0.6}$ 	&$0.20 $ & $0$  \\
			~ & \textbf{B4} & ${0.6}$ 	&$0.4 $ & $0$  \\
			~ & \textbf{B5} & ${0.1}$ 	&$0.9 $ & $0$  \\
			~ & \textbf{B6} & ${0.5}$ 	&$0.5 $ & $0$  \\
			~ & \textbf{B7} & ${1}$ 	&$0 $ & $0$  \\
			~ & \textbf{B8} & ${1}$ 	&$0 $ & $0$  \\
			~ & \textbf{B9} & ${0.3}$ 	&$0.7 $ & $0$  \\
			~ & \textbf{B10} & ${0.6}$ 	&$0.4 $ & $0$  \\
			~ & \textbf{B11} & ${0.5}$ 	&$0.4 $ & $0.1$  \\
			~ & \textbf{B12} & ${1}$ 	&$0 $ & $0$ \\
			~ & \textbf{B13} & ${1}$ 	&$0 $ & $0$\\
			~ & \textbf{B14} & ${0.6}$ 	&$0.4 $ & $0$  \\
			~ & \textbf{B15} & ${0.6}$ 	&$0.4 $ & $0$  \\
			~ & \textbf{B16} & ${0.4}$ 	&$0.6 $ & $0$  \\  \hline 			
			~ & \textbf{Avg.} & ${0.64}$ &$0.34 $ & $0.02$  \\  \hline 			
		\end{tabular}
		\vspace{2mm}
	\caption{Results achieved by the proposed method for each class in the MSRDA data set.}
	\label{tab:miscl_msrda}
	\end{scriptsize}
\end{table}

In Table~\ref{tab:comparison_msrda} we compare the average recognition rate achieved by the proposed method with the ones obtained by other existing approaches on the MSRDA data set. Note that a direct comparison is possible only with those methods that do not take into account information on the temporal sequence of poses. Other methods, instead, improve the classification performance and the inter-class robustness by including information about the temporal evolution of poses in the classification models. The method that obtains the highest value of the recognition rate ($68 \%$)  by only analyzing the skeleton information is based on Joint position~\cite{Wu12_dataset}. The results obtained by the proposed method are comparable to the ones reported in~\cite{Wu12_dataset}.

\begin{table}[!t]
	\begin{scriptsize}

\centering
\setlength\tabcolsep{0.2cm}
\renewcommand{\arraystretch}{1.3}
\begin{tabular}{|c|c|c|}
\hline 
\textbf{Method} & \textbf{Reference} & \textbf{Accuracy} \\
\hline  
LOP & \cite{Wu12_dataset} & $0.42$ \\
Depth Motion Maps & \cite{Yang12} & $0.43$ \\
Dynamic Temporal Warping & \cite{Muller06} & 0.54 \\
3DTSD & \cite{Koperski14} & 0.55 \\
Eigen Joints & \cite{Yang12_CVPR} & $0.58$ \\
TSD & \cite{Koperski14} & 0.58 \\
TSD+3DTSD & \cite{Koperski14} & 0.63 \\
\textbf{Proposed approach} & - & $\textbf{0.64}$ \\
TSD+RTD & \cite{Koperski14} & 0.65 \\
Joint Position & \cite{Wu12_dataset} & $0.68$ \\
Hierarchical Classification & \cite{Koperski14} & 0.72 \\
HON4D & \cite{Oreifej13} & 0.80 \\
Actionlet Ensamble & \cite{Wu12_dataset} & $0.85$ \\
\hline
\end{tabular}
\vspace{2mm}
\caption{Comparison of the results achieved by the proposed approach with the ones achieved by existing methods  on the MSRDA data set.}
	\label{tab:comparison_msrda}
		\end{scriptsize}
\end{table}

One advantage of the proposed approach is its \emph{trainability}. The selectivity of the pose detectors is not fixed in the implementation, but it is rather learned during an automatic configuration process, which is carried out on prototype patterns. 
In order to configure a limited set of detectors for a given action, we sampled the skeleton pose sequence by selecting only a confined number of poses which we used as prototypes for the configuration of a number of pose estimators. 

They key aspect of this approach is that the features are not designed by the user but are learned from training samples by means of an automatic configuration procedure. Hence, trainable features avoid a step of feature engineering, which usually require extensive domain knowledge to construct a set of hand-crafted features. The automatic configuration of features for pose detection that is part of the proposed method can be considered a kind of \emph{representation learning} that, similar to deep learning approaches, construct suitable representations from training samples.
In contrast with deep learning methods, the proposed trainable pose detectors do not require large amount of data to be configured. A single detector require only one training sample. However, the proposed approach learns representations that are less general than the one learned with deep learning.

Most of the classification errors are attributable to impossibility of distinguishing action due to the lack of information about the temporal sequence of poses. In further works, the proposed pose detection method can be employed as basic descriptor and coupled with a methodology for analysis of temporal sequences. In this way, a higher level classification approach would integrate on a larger time scale the classifications performed at frame-level.
An interesting way of improving the performance of the proposed method is to reduce the number of configured detectors and select only the most relevant ones, i.e. feature selection procedures can be applied as shown in similar works on selection of relevant trainable features~\cite{Strisciuglio15,Strisciuglio2016}. Reducing the number of configured detectors determines less computation resources and allows to configure a system for the detection of a larger number of classes.
Further improvements of the proposed method concern an extensive sensitivity analysis, which is meant to asses the robustness of the proposed approach with respect to the values of its  parameters. Furthermore, at the light of the high discriminant capability on the skeleton data demonstrated by the proposed approach, a combination with other typologies of descriptors using different information (for instance extracted by the depth images) as well as the introduction of the temporal information, will surely allow to deal in a more effective way with the problem of recognizing human actions.


\section{Conclusions}
\label{sec:concl}

In this paper we proposed a trainable feature detector for the recognition of human poses based on the analysis of the skeleton.   
The automatic configuration process of the proposed approach can be considered a kind of \emph{representation learning}, where the features are learned directly from training data instead of being had-crafted by an expert. 

The proposed detectors are able to evaluate the spatial arrangement of the joints of a skeleton in comparison with a given a prototype pattern of interest. The similarity between a prototype skeleton and a test skeleton is measured by taking into account some tolerance in the relative positions of the joints. This allows for generalization capabilities and robustness to distortion and noise.

The results that we achieved on a publicly available data set (average recognition rate of $64\%$ on the MSRDA data set) confirm the effectiveness of the proposed method and are comparable with the ones obtained by existing methods based on skeleton analysis.

\bibliographystyle{splncs03}
\bibliography{reacts}
\end{document}